\documentclass[10pt,twocolumn,letterpaper]{article}

\usepackage{acpr}
\usepackage{times}
\usepackage{epsfig}
\usepackage{graphicx}
\usepackage{amsmath}
\usepackage{amssymb}
\usepackage{caption}
\usepackage{blindtext}
\usepackage{cite}

\usepackage[pagebackref=true,breaklinks=true,letterpaper=true,colorlinks,bookmarks=false]{hyperref}

\acprfinalcopy 

\ifacprfinal\fi

\begin{document}

\title{
Style Transfer for Anime Sketches\\ 
with Enhanced Residual U-net and Auxiliary Classifier GAN
}
\author{Lvmin Zhang, Yi Ji and Xin Lin\\
School of Computer Science and Technology, Soochow University\\
Suzhou, China\\
{\tt\small lmzhang9@stu.suda.edu.cn, jiyi@suda.edu.cn}
}
\twocolumn[{%
\renewcommand\twocolumn[1][]{#1}%
\maketitle
\begin{center}
    \centering
    \includegraphics[width=\textwidth]{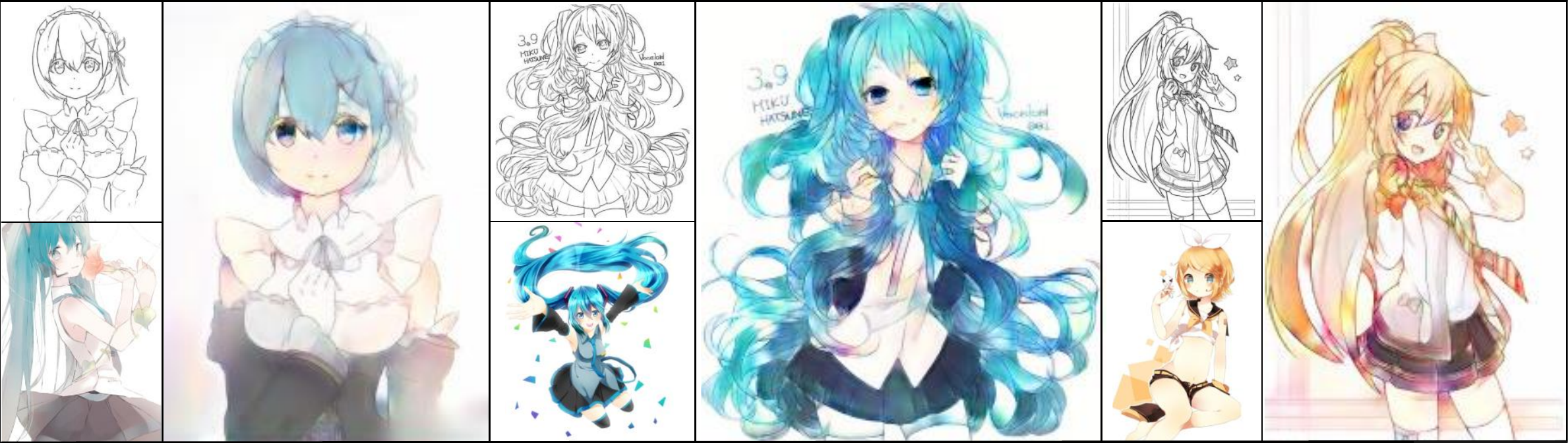}
    \captionof{figure}{Examples of combination results on sketch images (top-left) and style images (bottom-left). Our approach automatically applies the semantic features of an existing painting to an unfinished sketch. Our network has learned to classify the hair, eyes, skin and clothes, and has the ability to paint these features according to a sketch. More results can be seen at the end of paper.}\label{fig:tit}
\end{center}%
}]


\begin{abstract}
   Recently, with the revolutionary neural style transferring methods\cite{Gatys2015A,Gatys2016Image,Gatys2016Preserving,Johnson2016Perceptual,Ulyanov2016Texture}, creditable paintings can be synthesized automatically from content images and style images. However, when it comes to the task of applying a painting's style to a anime sketch, these methods will just randomly colorize sketch lines as outputs (fig.~\ref{fig:fail}) and fail in the main task: specific style tranfer. In this paper, we integrated residual U-net to apply the style to the grayscale sketch with auxiliary classifier generative adversarial network (AC-GAN)\cite{Odena2017Conditional}. The whole process is automatic and fast, and the results are creditable in the quality of art style as well as colorization.
\end{abstract}

\begin{figure*}
\begin{center}
\includegraphics[width=\textwidth]{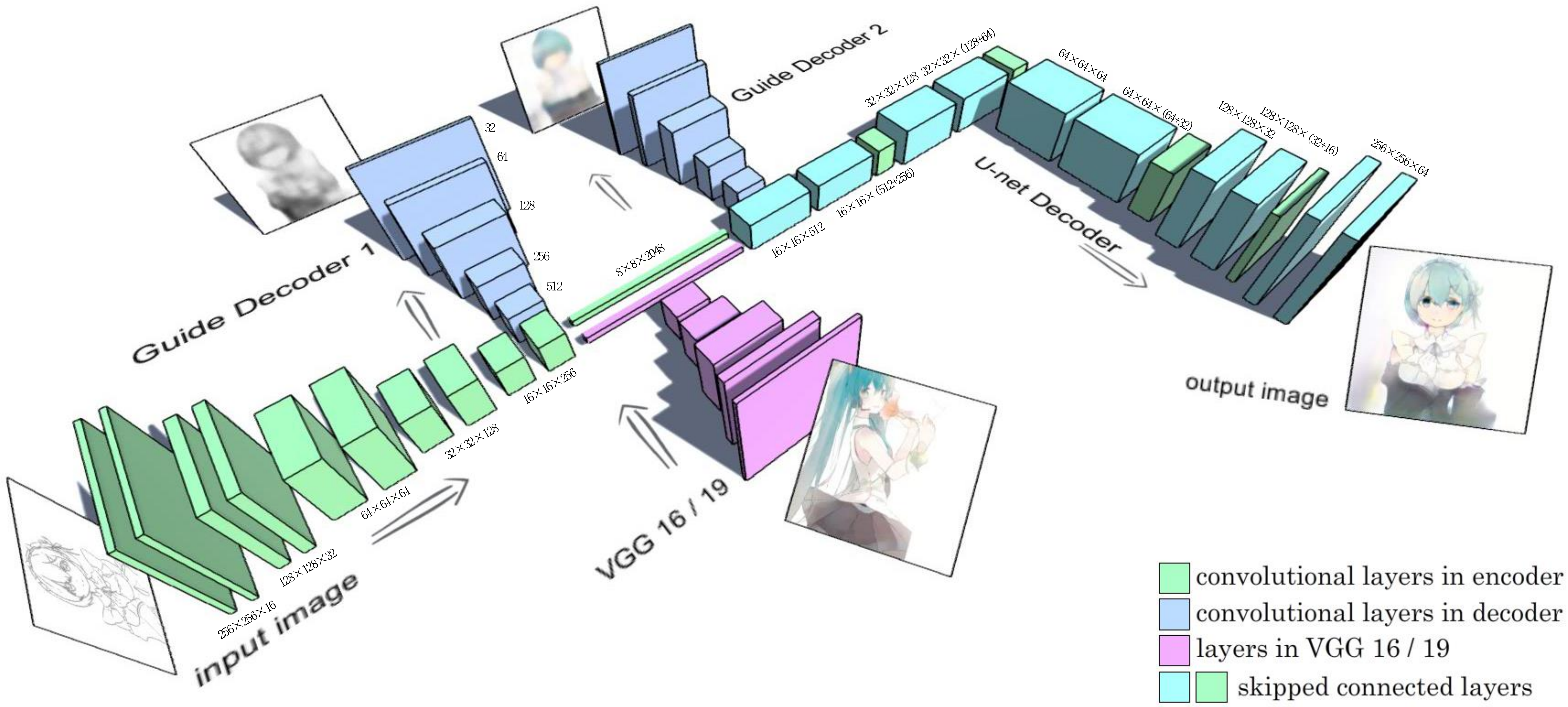}
\end{center}
   \caption{The architecture of the generator. The generator is based on U-net with skip connected layers. The weights of the VGG 16 or 19 \cite{Simonyan2014Very} is locked in order to avoid being disturbed while training. The two "Guide Decoder" are located at the entry and exit of mid-level layers. The 4096 outputs of VGG's \emph{fc1} are regarded as global style hint and added to the mid-level layers after a global normalization. For better performance in training, we add a dense 2048 layer above the \emph{fc1}.}
\label{fig:generator}
\end{figure*}

\section{Introduction}
In the community of animation, comics and games, the majority of artistic works can be created from sketches, which consumed a lot of time and effort. \textbf{If there is a method to apply the style of a painting to a half-finished work in form of sketches, many redundant work will be saved}, e.g. using an existing picture of a specific character as a style map and then apply the style to a sketch of the character. The neural algorithm of artistic style \cite{Gatys2015A} can produce amazing and perfect images combining of content maps and style maps, but it lacks the ability to deal with sketches. The paintschainer \cite{pc}, as well as pix2pix \cite{Isola2016Image}, can turn sketches into paintings directly and the result can be even perfect with pointed hints added, but it cannot take the advantage of existing paintings. We investigated residual U-net and auxiliary classifier GAN (AC-GAN) as a solution and our network can directly generate a combination of sketch and style image. Our model is fully feed-forward so as to synthetize paintings at a high speed. Furthermore, we find out that U-net and conditional GAN (cGAN) relatively declines in performance with absence of a balanced quantity of information of paired input and output, and we propose using residual U-net with two guide decoders. In addition, we compare many kinds of GANs and find that conditional GAN is not suitable for this task, resorting to the AC-GAN finally.

\textbf{Our contributions are}:
\begin{itemize}
\item{A feed-forward network to apply the style of a painting to a sketch.}
\item{An enhanced residual U-net capable of handling paired images with unbalanced information quantity.}
\item{An effective way to train residual U-net with two additional loss.}
\item{A discriminator modified from AC-GAN suitable to deal with paintings of different style.}
\end{itemize}

\section{Related Works}

\textbf{Neural Style Transfer} \cite{Gatys2015A,Gatys2016Image,Gatys2016Preserving,Johnson2016Perceptual,Ulyanov2016Texture} can synthesize admirable image with art style from an image and content from another, based on an algorithm that minimize the difference in gram matrixes of deep convolution layers. Nevertheless, our objective is to combine a style image and a sketch. Unfortunately, the neural style transfer is not capable of this kind of task. In fact, the results of neural style transfer on sketches from style images can be really strange, far from a proper painting.

\textbf{Pix2Pix} \cite{Isola2016Image} and some other paired image2image transfers based on conditional GAN \cite{Mirza2014Conditional} are accomplished in transformation between paired images. In our research, the quality of the outputs of networks based on cGAN depends on the “information gap degree” between the inputs and outputs. That is to say, if the input and output are similar in the quantity of information, the result can be reliable. In the experiment of Pix2Pix's edge2cat, based on users' input edges, small circles always means eyes, triangles regarded as ears, and closed figures should always be filled with cat hair. If we shuffle the datasets of cat2edge, bag2edge, and even more like house, bicycle, tree, dog and so on, the quality of outputs will decline accordingly. Our task is to transfer sketches correspondingly to paintings, which is far more complicated than cats or bags, and the generator of cGAN needs to learn semantic features and low-level features simultaneously. Actually, a conditional discriminator can easily leads the generator to focus too much on the relationship between sketches and paintings, thus, to some extent, ignore the composition of a painting, leading to ineluctable overfitting.

\textbf{Paintschainer} \cite{pc} has abandoned cGAN and resorted to an unconditional discriminator due to the same reason as above, and obtained remarkable and impressive achievements. It becomes a powerful and popular tool for painters. Users only need to input a sketch to get a colorized painting. They can add pointed color hints for better results. With massive existing finished paintings, though the demand for a method to colorize a sketch according to a specific style is extremly high, there is no reliable and mature solution for it. 

\section{Methods}

We combine an enhanced residual U-net generator and an auxiliary classifier discriminator as our adversarial network. We feed the generator with sketch maps and style maps. Our discriminator can tell whether its input is real or fake, and classify corresponding style simultaneously.

\begin{figure}
\centering\includegraphics[width=.5\textwidth]{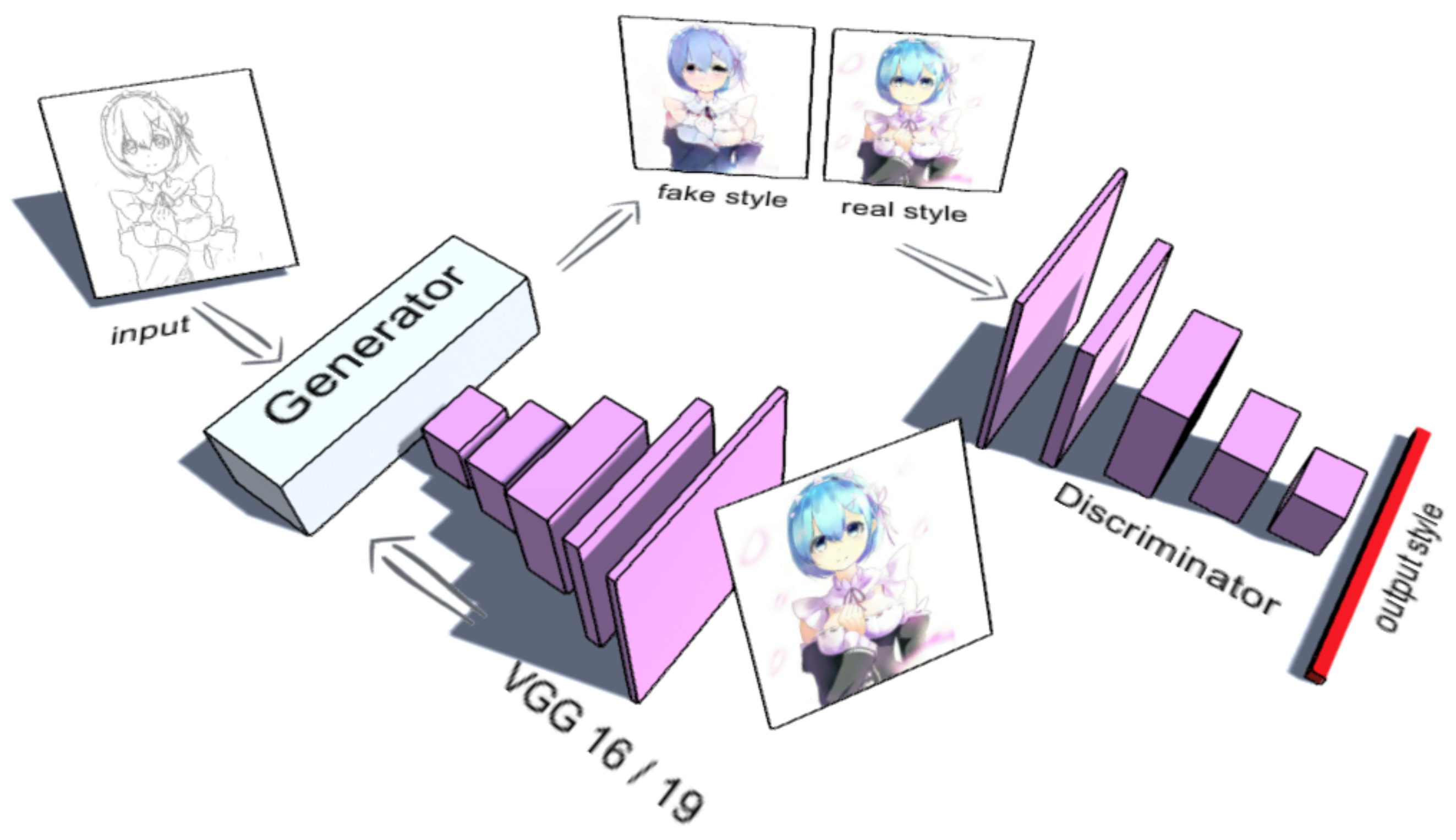}
\caption{The architecture of our adversarial network. The discriminator is modified from AC-GAN, which not only has the ability to reveal whether the map is real or fake, but also tell the classification. We notice that the global style hint can be regarded as a low-level classification result with 2048 or 4096 classes.}\label{fig:discriminator}
\end{figure}

\begin{figure}
\centering\includegraphics[width=.35\textwidth]{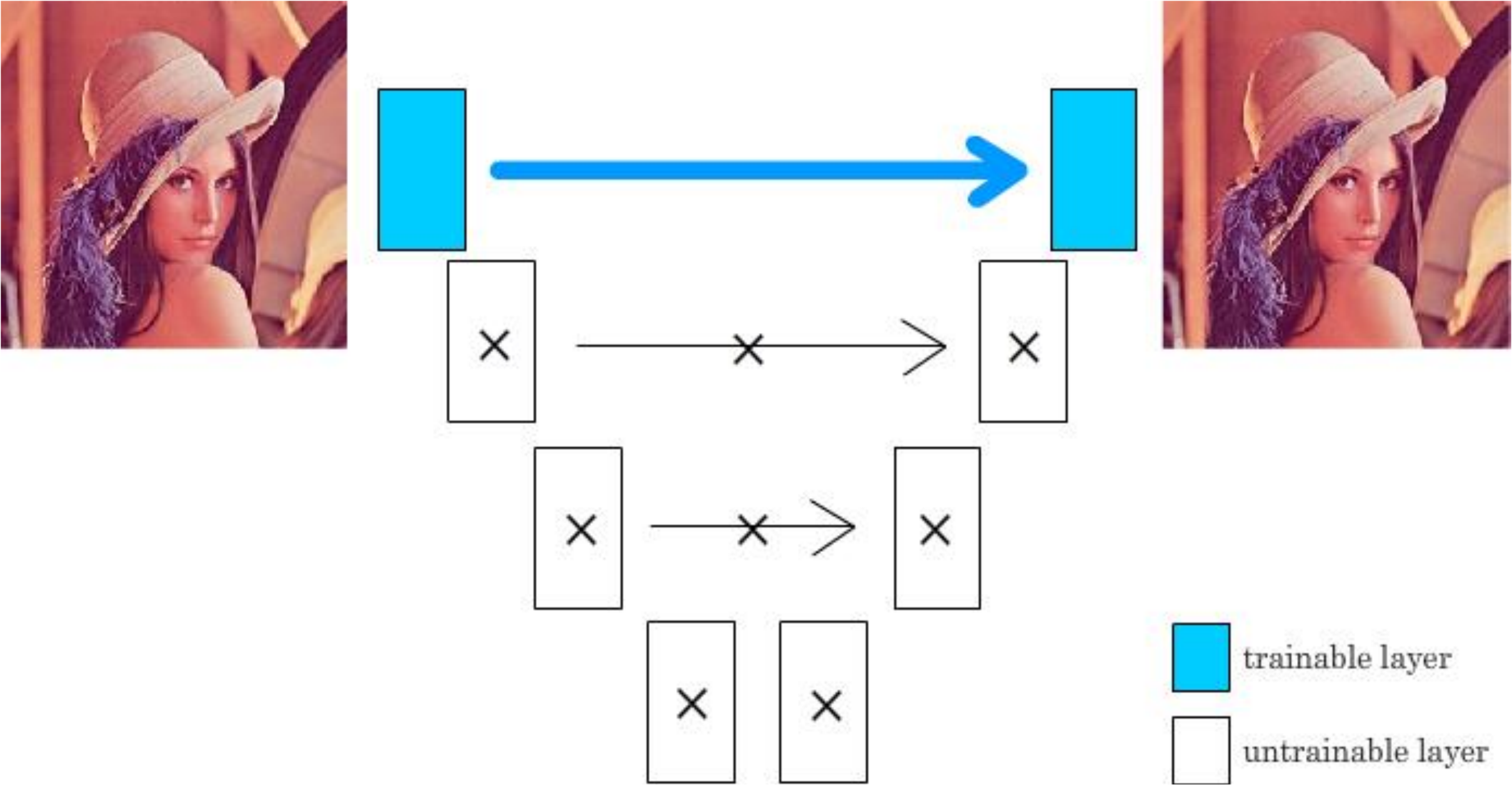}
\caption{In the experiment of copying images, the mid-level layers of U-net receives no gradients.}\label{fig:copy}
\end{figure}

\subsection{Architecture and Objective of Generator}
\begin{figure}
\centering\includegraphics[width=.48\textwidth]{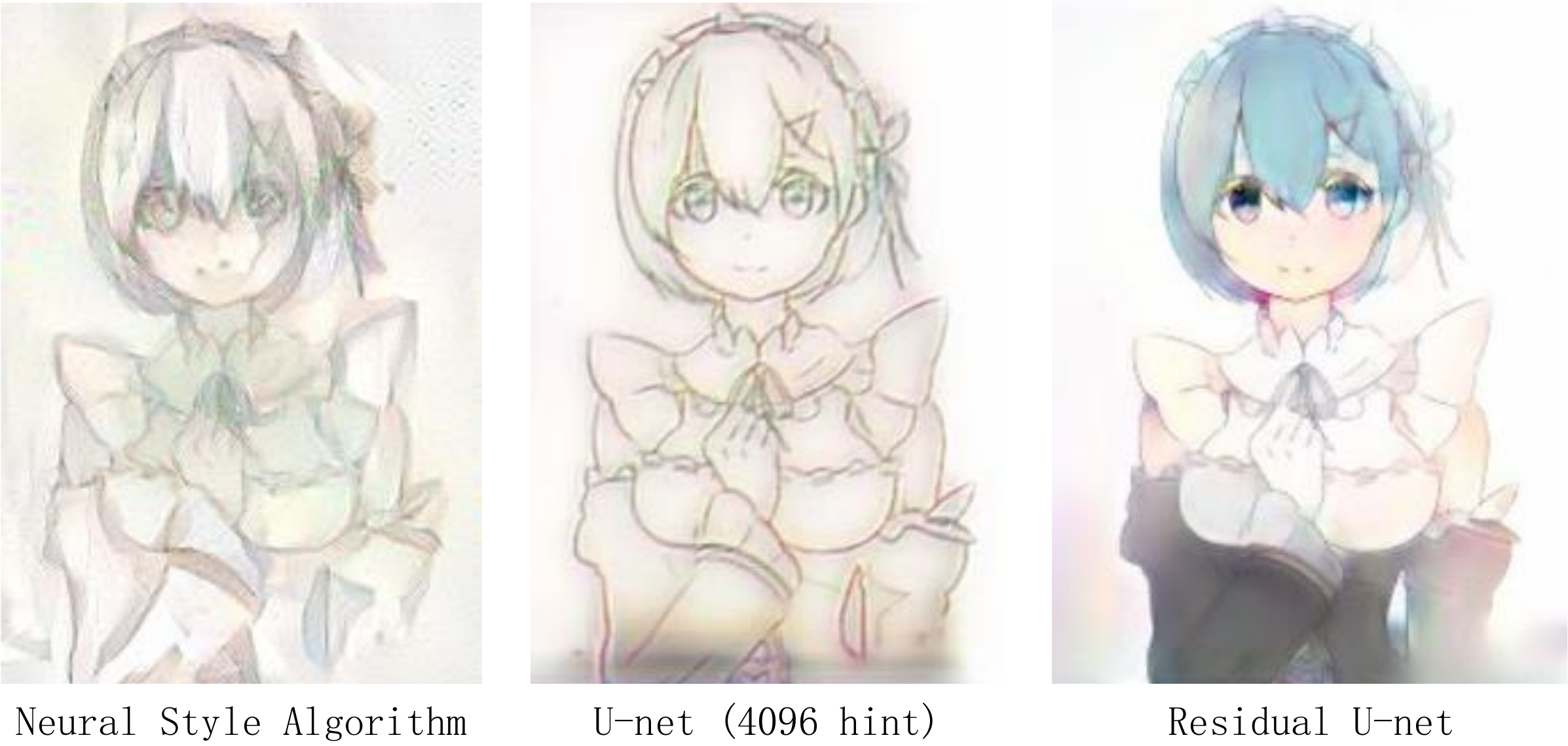}
\caption{The style map and sketch map are same with fig.~\ref{fig:tit}. The neural style algorithm fails to transfer a sketch to a painting. Though normal U-net can predict meaningful paintings when the global hint is 1$\times$1$\times$256, the result can be rather disappointing when the hint of 1$\times$1$\times$4096 is applied. Our residual U-net with two Guide Decoders is very capable of handling such an informative global style hint.}\label{fig:fail}
\end{figure}

The detailed structure is in fig.~\ref{fig:generator}. We are not the first to add the global style hint to mid-level layers of a encoder-decoder or a U-net\cite{Ronneberger2015U}. In \emph{Iizuka et.al's} \cite{Iizuka2016Let} and \emph{Richard Zhang et.al's} \cite{Zhang2017Real} the global hint is extracted from a high-level layer of a classification network as a high dimension vector and then added to the colorization network. For photograph colorization, the shadow, material and texture is known variables in inputs and the network only need a spot of information to analyse the color distribution. In \emph{Iizuka et.al's} \cite{Iizuka2016Let}, they only use vectors of 1$\times$1$\times$256 as global hints. However, it is far from enough for the network to paint sketches into paintings with various unique styles. Therefore, We integrate style hint at size of 1$\times$1$\times$4096 or 1$\times$1$\times$2048. But we failed to train such a generator directly and finally found some "nature" of the well-known structure U-net. The failure result is in fig.~\ref{fig:fail}.

\textbf{The U-net is "lazy".} That is to say if the U-net find itself able to handle a problem in low-level layers, the high-level layers will not bother to learn anything. If we train a U-net to do a very simple work "copying image" as in fig.~\ref{fig:copy}, where the inputs and outputs are same, the loss value will drop to 0 immediately. Because the first layer of encoder discovers that it can simply transmit all features directly to the last layer of the decoder by skiping connection to minimize the loss. In this case, no matter how many times we train the U-net, the mid-level layers will not get any gradient.

For each layer in U-net's decoder, features can be acquired from higher layers or from skip connected layers. In each iteration of a training process, these layers select other layers' outputs with nonlinear activations in order to minimize the loss. In the experiment of copying image, when the U-net is initialized with Gaussian random numbers, the output of the first layer in encoder is informative enough to express the full input map while the output of second-to-last layer in decoder seems very noisy. Thus the "lazy" U-net gives up the relatively noisy features. 

\begin{figure}
\centering\includegraphics[width=.35\textwidth]{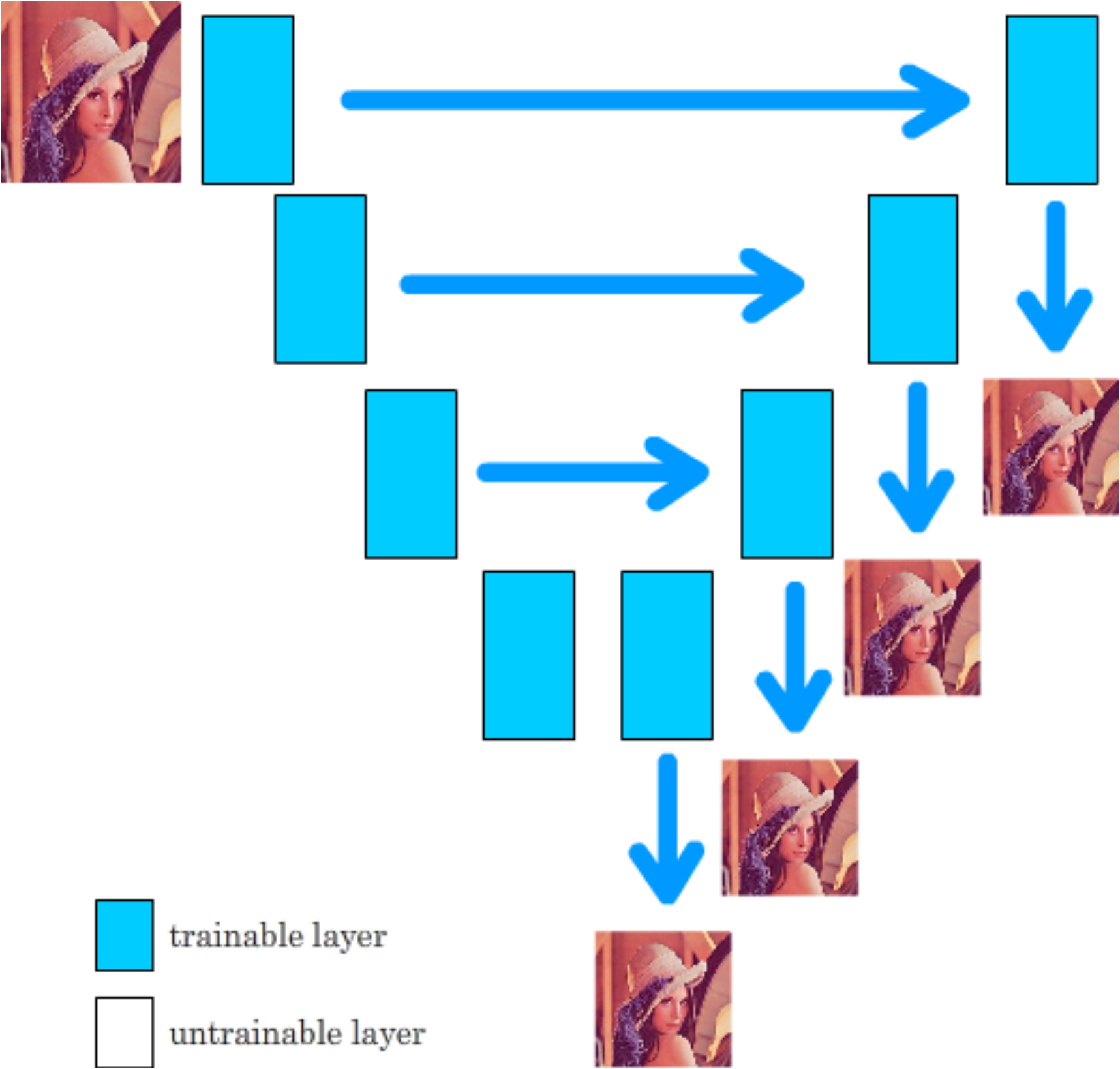}
\caption{The ideal architecture of a residual U-net. Additional losses are applied to all mid-level layers so as to provide gradients for these layers.}\label{fig:iealunet}
\end{figure}

Because a hint of 1$\times$1$\times$256 is far from enough for sketch painting, we resorted to a hint of 1$\times$1$\times$4096, from the output of VGG 19's \emph{fc1}, without the \emph{Relu} activation. The 4096 hint vector is very informative and powerful. However, for a newly initialized U-net, the output of the mid-level layers can be extremely noisy if we add the vector of 4096 directly to the layer. As mentioned above, the noisy mid-level layer is given up by U-net and these layers cannot receive any gradient as a consequence.

\begin{figure}
\centering\includegraphics[width=.5\textwidth]{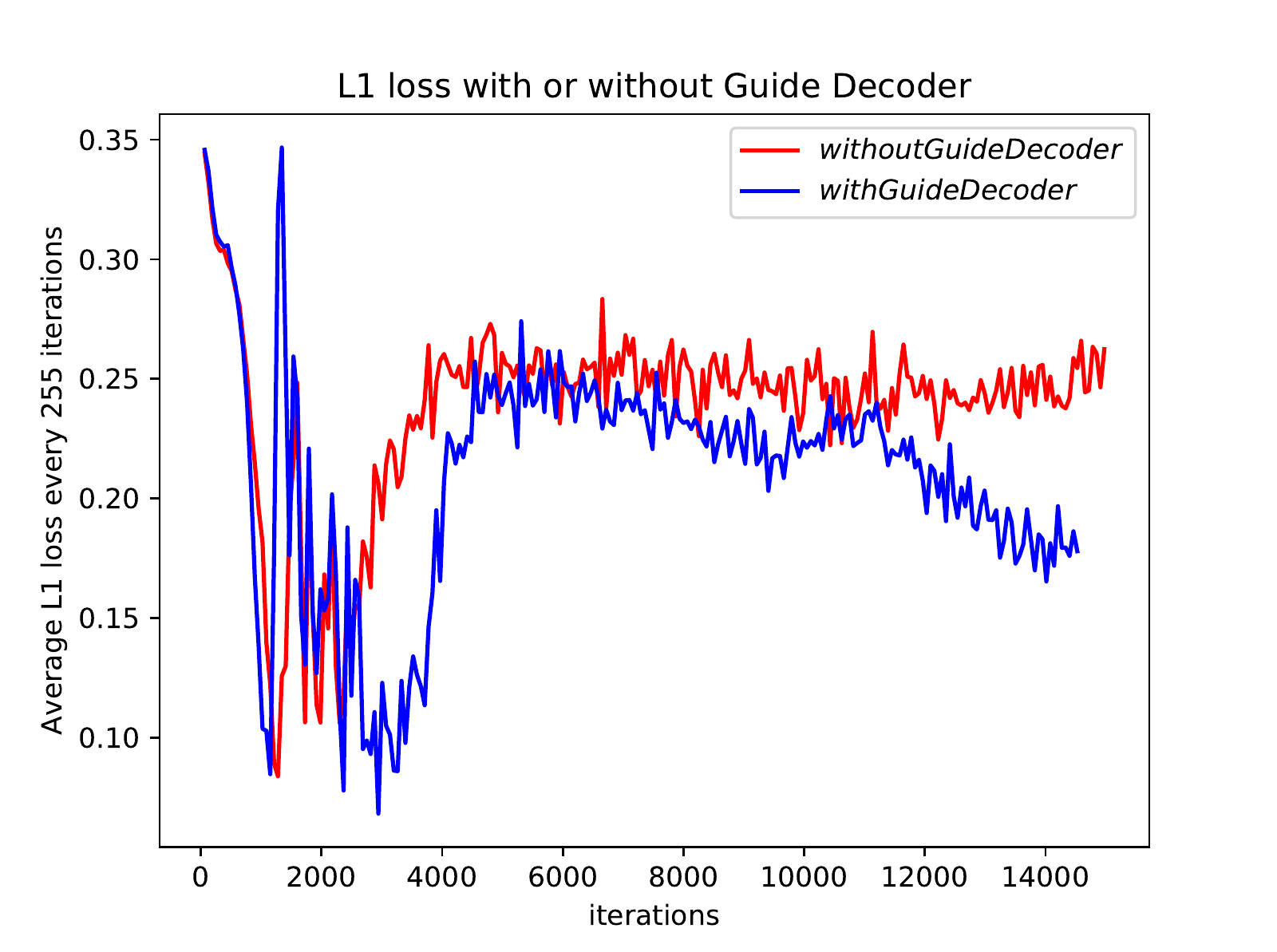}
\caption{Loss with or without the Guide Decoder}\label{fig:wo}
\end{figure}

Inspired by LeNet \cite{LecunLeNet} and GooLeNet \cite{Szegedy2014Going}, We resort to residual networks in fig.~\ref{fig:iealunet}. If we attach additional loss to each layer which is possible to be "lazy", no matter how noisy the output of a mid-level layer is, the layer will never be given up by the U-net and all layers will get stable gradient in the whole training process. Thus, it is possible to add a very informative and to some extent noisy hint to the mid-level layers. We implemented two additional loss, in the "Guide decoder 1" and "Guide decoder 2", to avoid the gradient disappearance in mid-level layers. The loss trend with or without the two Guide Decoders can been seen in fig.~\ref{fig:wo}. Significant difference of the networks' prediction can be seen in fig.~\ref{fig:fail}.

The loss is defined as:
\begin{equation}
\label{eqn3_2}
\begin{aligned}
L_{l1}(V,G_{f,g_{1},g_{2}})=\mathbb{E}_{x,y\sim P_{data}(x,y)}[||y-G_{f}(x,V(x))||_{1}+\\\alpha||y-G_{g_{1}}(x)||_{1}+\beta||y-G_{g_{2}}(x,V(x))||_{1}]
\end{aligned}
\end{equation}

Where the $x$, $y$ is the paired domain of sketches and paintings, and $V(x)$ is the output of VGG 19's $fc1$ without $Relu$, and $G_{f}(x,V(x))$ is the final output of U-net, the $G_{g_{1}}(x)$ and $G_{g_{2}}(x,V(x))$ are outputs of the two guide decoders at the entry and the exit of mid-level layers accordingly. The recommended value of $\alpha$ and $\beta$ is \emph{0.3} and \emph{0.9}. We also find out that the distribution of color can be improved by feeding the Guide Decoder located at the entry of mid-level layers with grayscale maps, so the final loss is as below, where $T(y)$ can transform y into grayscale image.
\begin{equation}
\label{eqn3_2}
\begin{aligned}
L_{l1}(V,G_{f,g_{1},g_{2}})=\mathbb{E}_{x,y\sim P_{data}(x,y)}[||y-G_{f}(x,V(x))||_{1}+\\\alpha||T(y)-G_{g_{1}}(x)||_{1}+\beta||y-G_{g_{2}}(x,V(x))||_{1}]
\end{aligned}
\end{equation}

\subsection{Architecture and Objective of Discriminator}

As we mentioned, the traditional cGAN is not suitable for our project. Painting is a complicated work and needs human artists to take color-selection, composition and fine-tuning into consideration, and all these need an artist to focus on the global manner of the painting. However, a conditional discriminator has always a tendency to focus much more on the relationship between sketch line and color than the global information. In our experiments with Pix2Pix, if a conditional discriminator is applied, the generator will resist fiercely and the color surrounding the line can be extremely over-colorized. Tuning parameters is not enough to reach a balance.

Furthermore, it can be appreciated if the discriminator has the ability to tell the style and provide gradient accordingly, in order to fit the main task: style transfering. We finally integrate AC-GAN and our discriminator has exactly 4096 outputs, which will all be minimized to 0 when the input is fake and approach to the same value of VGG 19's \emph{fc1} when the input is real in fig.~\ref{fig:discriminator}. 

The final loss is defined as:
\begin{equation}
\label{eqn3_2}
\begin{aligned}
L_{GAN}(V,G_{f},D)=\mathbb{E}_{y\sim P_{data}(y)}[Log(D(y)+(1-V(y)))]+\\\mathbb{E}_{x\sim P_{data}(x)}[Log(1-D(G_{f}(x,V(x))))]
\end{aligned}
\end{equation}

Unfortunately, to minimize the loss, it requires a large memory size in GPU. We also employed the traditional discriminator of DCGAN for fast speed and large batch size. We fine-tune our networks by shifting the two loss functions.
\begin{equation}
\label{eqn3_2}
\begin{aligned}
L_{GAN}(V,G_{f},D)=\mathbb{E}_{y\sim P_{data}(y)}[Log(D(y))]+\\\mathbb{E}_{x\sim P_{data}(x)}[Log(1-D(G_{f}(x,V(x))))]
\end{aligned}
\end{equation}

The final objective is:
\begin{equation}
\label{eqn3_2}
\begin{aligned}
G^{*}= \arg \min \limits_{G_{f}} \max \limits_{D}L_{GAN}(V,G_{f},D) + \lambda L_{l1}(V,G_{f,g_{1},g_{2}})
\end{aligned}
\end{equation}

\begin{figure*}
\centering\includegraphics[width=\textwidth]{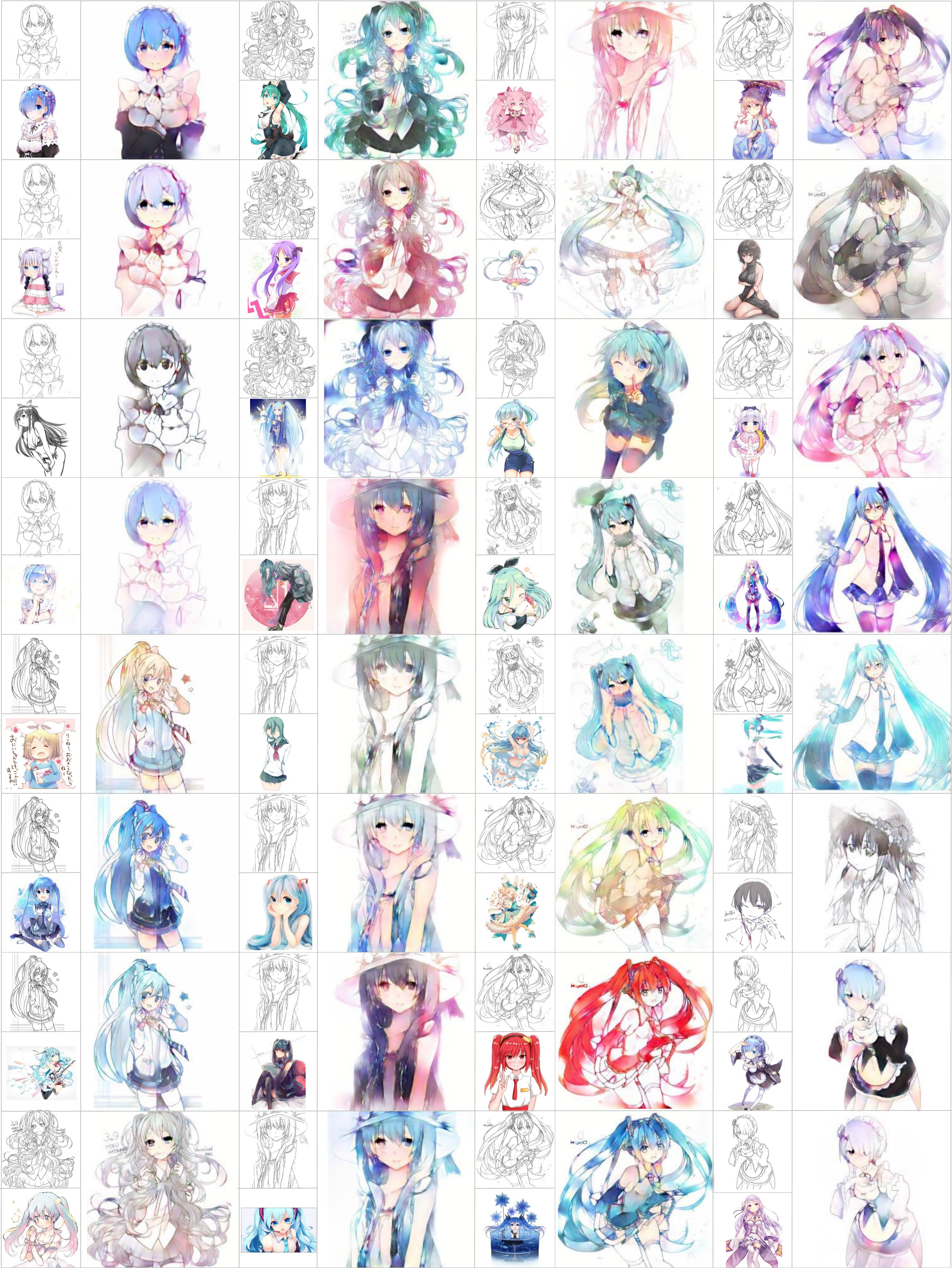}
\caption{Additional results of our network. All the reference paintings are from Pixiv. The creditions for all artists are available in the github page "style2paints".}\label{fig:4}
\end{figure*}

\section{Limitations and Discussions}

Our network is capable of drawing on sketches according to style maps given by users, depending on the classification ability of the VGG. However, the pretrained VGG is for ImageNet photograph classification, but not for paintings. In the future, we will train a classification network only for paintings to achieve better results. Furthermore, due to the large quantity of layers in our residual network, the batch size during training is limited to no more than 4. It remains for future study to reach a balance between the batch size and quantity of layers.

\nocite{*}

\bibliographystyle{plain}%
\bibliography{sketch_transfer}

\end{document}